%% file: main.tex
\documentclass[10pt,twocolumn,letterpaper]{article}

\usepackage{times}
\usepackage{epsfig}
\usepackage{graphicx}
\usepackage{amsmath}
\usepackage{amssymb}
\usepackage{lipsum}
\usepackage{adjustbox}

\usepackage{color}
\usepackage{colortbl}

\usepackage{enumitem}
\usepackage{booktabs}
\usepackage{bm}
\usepackage{wrapfig}
\usepackage{cite}
\usepackage{algorithm}
\usepackage{booktabs}
\usepackage{graphicx}
\usepackage[noend]{algpseudocode}
\usepackage{lipsum}

\usepackage{makecell}
\usepackage{multirow}

\usepackage{subcaption}
\usepackage{xspace}

\usepackage{pifont}
\usepackage{bm}
\usepackage{bbding}

\usepackage{cite}
\usepackage{cvpr}              

\input{preamble}

\definecolor{cvprblue}{rgb}{0.21,0.49,0.74}
\def\ie{\textit{i.e.}} 
\def\vs{\textit{vs.}} 
\def\eg{\textit{e.g.}} 
\usepackage[pagebackref,breaklinks,colorlinks,citecolor=cvprblue]{hyperref}

\newcommand{\ourmethod}{MixCon3D}

\newcommand{\yp}[1]{\textcolor[rgb]{0.0,0.0,0.0}{\@#1}}

\newcommand{\blue}[1]{\textcolor[rgb]{0.0,0.0,0.9}{\@#1}}

\def\paperID{8626} 
\def\confName{CVPR}
\def\confYear{2024}

\title{Sculpting Holistic 3D Representation \\ in Contrastive Language-Image-3D Pre-training}

\author{Yipeng Gao$^{1}$ \qquad Zeyu Wang$^{2}$ \qquad Wei-Shi Zheng$^{1}$ \qquad Cihang Xie$^2$ \qquad Yuyin Zhou$^{2}$ \\
\vspace{12pt}
$^1$Sun Yat-sen University \qquad $^2$University of California, Santa Cruz\\} 

\begin{document}
\maketitle
\begin{abstract}
Contrastive learning has emerged as a promising paradigm for 3D open-world understanding, \ie, aligning point cloud representation to image and text embedding space individually.
In this paper, we introduce \ourmethod, \yp{a simple yet effective method aiming to sculpt holistic 3D representation in contrastive language-image-3D pre-training.}
\yp{In contrast to point cloud only, we develop the 3D object-level representation from complementary perspectives, \eg, multi-view rendered images with the point cloud.}
\yp{Then, \ourmethod~performs language-3D contrastive learning, comprehensively depicting real-world 3D objects and bolstering text alignment.}
Additionally, we pioneer the first thorough investigation of various training recipes for the 3D contrastive learning paradigm, building a solid baseline with improved performance.
Extensive experiments conducted on three representative benchmarks reveal that our method significantly improves over the baseline, surpassing the previous state-of-the-art performance on the challenging 1,156-category Objaverse-LVIS dataset by \textbf{5.7\%}. 
\yp{The versatility of \ourmethod~is showcased in applications such as text-to-3D retrieval and point cloud captioning, further evidencing its efficacy in diverse scenarios.}
The code is available at \href{https://github.com/UCSC-VLAA/MixCon3D}{https://github.com/UCSC-VLAA/MixCon3D}. 
\end{abstract}

\section{Introduction}
\label{sec:intro}
The ability to perceive and comprehend 3D environments is crucial in applications like augmented and virtual reality, autonomous driving, and embodied AI.
Despite significant progress achieved in closed-set 3D recognition \cite{pu-gcn, dynGCN, gcn_attn, pointnet, pointnet++, pointnext, point-bert, stratified_transformer, point_transformer}, there is still a distinct gap between the advanced development of 2D and 3D vision methods.
This phenomenon primarily stems from the limited diversity and complexity of existing 3D datasets caused by high data acquisition costs. 

Recent research endeavors have turned to well-trained 2D foundation models to unlock the full potential of 3D open-world recognition.
A line of such works is built upon CLIP~\cite{openai_clip}, a pioneering foundation model known for its extraordinary zero-shot recognition capability~\cite{clip_video_zero_shot, clip_video_zero_shot_2, clip_OVOD, clip_OVOD_seg, clip_cls} by training on web-scale data~\cite{laion400m, laion5b}. 
The knowledge learned from millions or even billions of image-text pairs proves to be invaluable in assisting the model to learn 3D shapes. 
In this context, ULIP~\cite{ulip} and CLIP$^{2}$~\cite{zeng2023clip2} first propose to keep the image and text encoder frozen while training the 3D encoder on the (image, text, point cloud) triplets, which leads to substantially increased zero-shot 3D recognition performance. 
\begin{figure}[!t]
  \centering
  \includegraphics[width=0.9\linewidth]{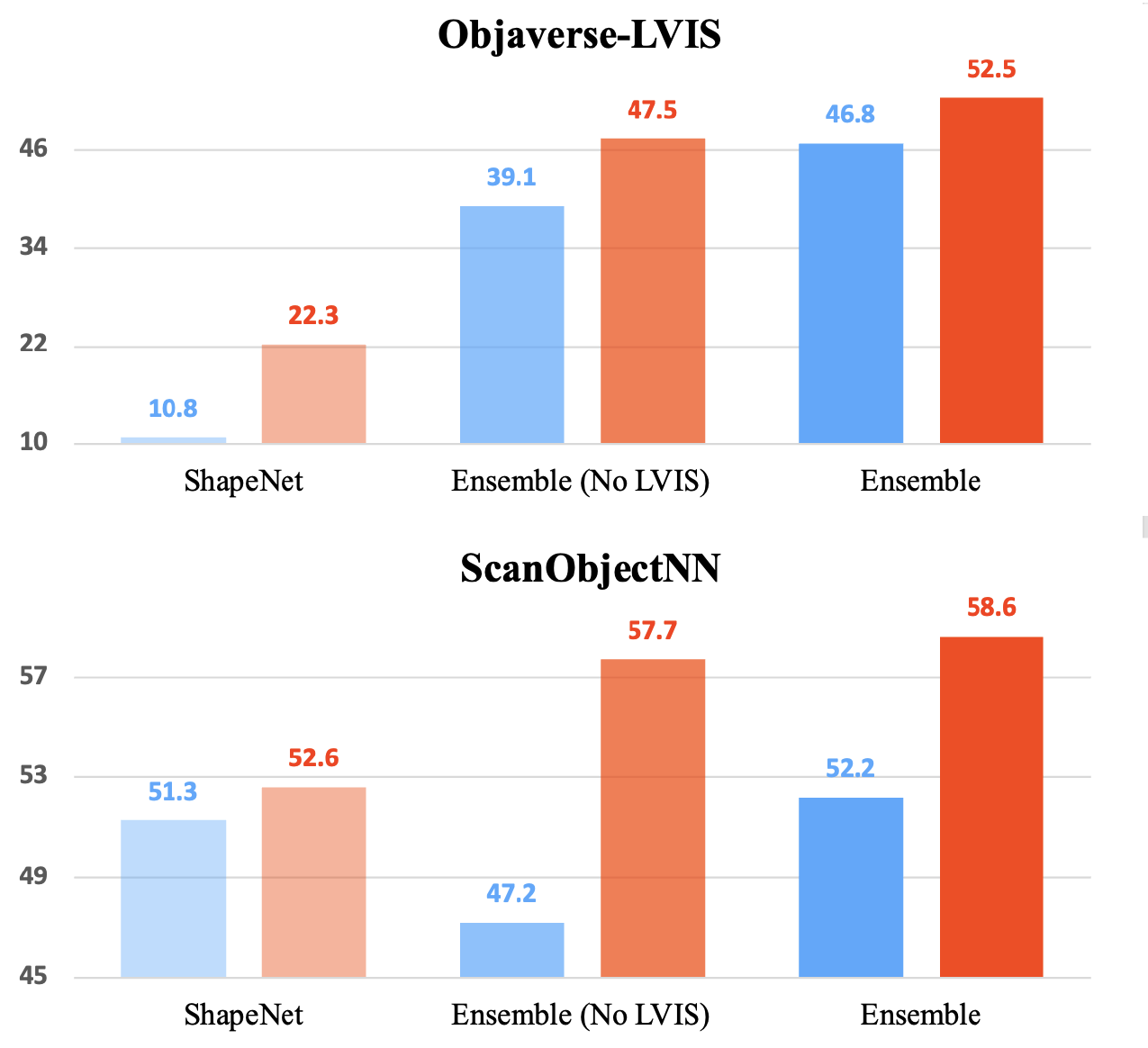}
  \vspace{-1em}
   \caption{Comparison of zero-shot point cloud recognition between the OpenShape (\blue{blue}) and our \ourmethod~(\red{red}) under different pre-training datasets (ShapeNet, Ensemble (No LVIS) and Ensemble). 
   Our model obtains consistent improvements on different training datasets on various downstream benchmarks.}
   \vspace{-1em}
   \label{fig:intro}
\end{figure}
While existing methods have demonstrated great promise, they predominantly center on a vanilla correspondence between point-text and point-image to form contrastive pairs, typically overlooking the intricate relationships across various modalities and perspectives. 
For instance, multi-view RGB images and 3D point clouds are known to capture distinct yet complementary attributes of a 3D object~\cite{bai2022transfusion,liu2023bevfusion, chen2023bevdistill, Wang2023distillbev, MVConv,jaritz2019multi,hamdi2021mvtn} --- point clouds emphasize depth and geometry, whereas multi-view images excel at representing semantic information \yp{from diverse parts}. 
\yp{However, distinct characteristics from each modality of the same 3D object are isolated in the previous contrastive pre-training scheme.}

To bridge these gaps, in this paper, we propose to \yp{sculpt a comprehensive 3D object-level representation in contrastive language-image-3D pre-training}, termed as \textbf{\ourmethod}, a simple yet effective method tailored to maximize the efficacy and potential of contrastive learning across images, texts, and \yp{3D objects}.   
Central to our approach is \emph{utilizing the complementary information between multi-view 2D images and 3D point clouds to jointly represent a 3D object and align the \yp{3D object-level representation} to the text embedding space}.
Specifically, we construct holistic 3D representations by simply concatenating the point cloud features and fused multi-view projected image features before contrastive learning, thus fortifying cross-modal alignment.
We also establish an advanced training guideline by carefully examining the training recipe (\eg, batch size, temperature parameters, and learning rate schedules).
This not only stabilizes the training process but also drives enhanced performance. 

As illustrated in Figure~\ref{fig:intro}, our \ourmethod~consistently shows remarkable improvements over multiple popular 3D understanding benchmarks. 
For example, on the well-established ScanObjectNN dataset, our approach substantially outperforms the prior art by 6.4\%, demonstrating the strong generalization ability of \ourmethod. 
Moreover, on the challenging 1,156-category Objaverse-LVIS dataset with long-tailed distribution, our \ourmethod~attains an accuracy of 52.5\%, surpassing the competing models by a significant margin of 5.7\%. 
Lastly, by following OpenShape~\cite{openshape} to employ the learned 3D features in the tasks of text to 3D shape retrieval and point cloud caption generation, we showcase our newly learned 3D embedding space is well aligned with CLIP image and text embedding space. 

\section{Related Works}
\label{sec:related_works}
\paragraph{3D Representation Learning.}
Point-based methods, a prominent category of 3D representation learning, have garnered much attention for their simplicity, effectiveness, and efficiency. 
The pioneering work, PointNet~\cite{pointnet}, models the inherent permutation invariance of points with point-wise feature extraction and max-pooling, enabling direct processing of unstructured point sets.
PointNet++~\cite{pointnet++} enhances PointNet with a hierarchical network architecture to capture local and global geometric cues effectively. 
Building upon this foundation, the 3D community has witnessed the emergence of a plethora of point-based methods, with a particular focus on the design of effective local modules~\cite{pu-gcn,dynGCN,gcn_attn,KPconv,Tangent,spidercnn,relationconv,point_transformer}.
PointNext~\cite{pointnext} explores an orthogonal direction, underscoring the pivotal role of training and scaling strategies in effective 3D representation learning.

Another line of work focuses on designing self-supervised learning techniques tailored for point cloud understanding. 
Early endeavors along this direction centered around the proposition of various low-level pretext tasks, including self-reconstruction~\cite{achlioptas2018learning,deng2018ppf}, distortion reconstruction~\cite{sauder2019self,mersch2022self}, and normal estimation~\cite{rao2020global}.
Recently, the remarkable success of self-supervised learning in the language and vision domain has prompted researchers in the 3D domain to adopt analogous self-supervised learning paradigms~\cite{3D-L_pretrain, ponder}. 
PointContrast~\cite{xie2020pointcontrast}, for instance, leverages the concept of contrasting two views of the same point cloud to facilitate high-level scene understanding. 
PointBERT~\cite{point-bert} and PointMAE~\cite{pang2022masked}, based on the idea of masked modeling, train an autoencoder to recover the masked portion of data with the unmasked part of the input. 
Recent works~\cite{ACT_cross-modal, I2P_MAE, CrossPoint, simipu, P2P_cross_modal, pix4point} employ 2D models as an auxiliary tool to learn 3D point cloud representations.

Unlike designing 3D backbones or self-supervised learning pretext tasks, this paper focuses on multimodal contrastive learning for 3D open-world understanding.

\paragraph{CLIP for 3D open-world understanding.}
By training on web-scale image-text pairs, CLIP~\cite{openai_clip} has revolutionized the area of visual representation learning via language supervision. The extraordinary zero-shot recognition performance of CLIP has found applications in a lot of domains, including zero-shot text-to-3D generation~\cite{hong2022avatarclip,jain2022zero,michel2022text2mesh,sanghi2022clip, canfes2023text, aneja2023clipface,xu2023dream3d}, 
zero-shot 3D segmentation or detection~\cite{jatavallabhula2023conceptfusion,ding2023pla,yang2023regionplc,lu2023open, openscene, clip-fo3d},
and 3D shape understanding~\cite{zhang2022pointclip,zhu2022pointclipv2,ulip,ulip2,qi2023recon,clip_goes_3d}.
The early exploration in leveraging CLIP for 3D shape understanding typically involves the projection of the original point cloud into depth maps, followed by the direct application of 2D CLIP on them~\cite{zhang2022pointclip,zhu2022pointclipv2}. 
However, this approach suffers from information loss during projection while introducing extra latency.
Additionally, the domain gap between the synthetically rendered depth maps and natural images could significantly hurt CLIP performance. 

More recently, multiple works \cite{ulip,ulip2,openshape, vit-lens, Uni3D, TAMM} propose to learn a unified embedding space for image, text, and point cloud, through training a 3D encoder aligned with CLIP image/text encoder.
JM3D~\cite{JM3D} aligns the point cloud representation to a joint image-text embedding space, which ignores the completeness of 3D object representation.
In contrast to general modality fusion methods~\cite{GMC, MCLEA}, our work follows this line of work but takes one step ahead to reveal the power of constructing a more holistic 3D object representation.
We explore the importance of the complementarity between view difference and shape information for a comprehensive 3D object-level representation.

\section{\ourmethod}
\label{sec:review}

\subsection{Preliminaries}
\label{subsec:preliminary}
\paragraph{Optimization Objectives of Cross-modal Contrastive Learning.}
By exploiting a massive amount of image-text pairs crawled from the web, the CLIP model~\cite{openai_clip} has demonstrated exceptional open-world image understanding capability.
Typically, given batched image-text pairs $\{(\bm{x}_{i}^{I}, \bm{x}_{i}^{T})\}_{i=1}^{N}$ and the image, text encoders $f^{I}$, $f^{T}$, 
the CLIP is trained to bring the representations of paired image and text data $(\bm{x}_{i}^{I}, \bm{x}_{i}^{T})$ closer by the contrastive loss $\mathcal{L}^{I \leftrightarrow T}$ as follows:
\begin{equation}
\label{eq:logit1}
    l^{I \rightarrow T} = \sum_i^N\log\frac{\exp(\bm{z}_i^I\cdot \bm{z}_i^T/\tau)}{\sum_j\exp(\bm{z}_i^I\cdot \bm{z}_j^T/\tau)}
\end{equation}

\begin{equation}
\label{eq:logit2}
    l^{T \rightarrow I} = \sum_i^N\log\frac{\exp(\bm{z}_i^T\cdot \bm{z}_i^I/\tau)}{\sum_j \exp(\bm{z}_i^T\cdot \bm{z}_j^I/\tau)}
\end{equation}

\begin{equation}
\label{eq_clip_loss}
\begin{aligned}
\mathcal{L}^{I \leftrightarrow T}(\bm{x}_{i}^{I},\bm{x}_{i}^{T}) = -\frac{1}{2}(l^{I \rightarrow T} + l^{T \rightarrow I})
\end{aligned}
\end{equation}
where $\bm{z}^{I}_i = g^I \circ f^I(\bm{x}_{i}^{I})/||g^I \circ f^I(\bm{x}_{i}^{I})||$ and $\bm{z}^{T}_i = g^T \circ f^T(\bm{x}_{i}^{T})/||g^T \circ f^T(\bm{x}_{i}^{T})||$ are the $l_2$ normalized image and text features output by projection heads. 
$g^I$ and $g^T$ are image and text projection heads and $\tau$ is a learnable temperature.  

As the scale of 3D datasets is relatively smaller, previous works~\cite{ulip, ulip2, openshape, zeng2023clip2} have resorted to the pre-trained CLIP image and text embedding space for training a vanilla 3D model $g^P \circ f^{P}$ (including 3D encoder $f^P$ and projection head $g^P$) with open-world recognition ability. 
Since CLIP is pre-trained on a much larger data scale and is well aligned, its image model $g^I \circ f^{I}$ and text model $g^T \circ f^{T}$ are frozen during training.
Specifically, given batched $N$ input image $\bm{x}_{i}^{I}$, text $\bm{x}_{i}^{T}$, and point cloud $\bm{x}_{i}^{P}$ triplets $\{(\bm{x}_{i}^{I}, \bm{x}_{i}^{T}, \bm{x}_{i}^{P})\}_{i=1}^{N}$ (hence the name image-text-3D), the 3D model $g^P \circ f^{P}$ is trained to align the point cloud representation $\bm{z}^{P}_i = g^P \circ f^P(\bm{x}_{i}^{P})/||g^P \circ f^P(\bm{x}_{i}^{P})||$ to the CLIP embedding space by $\mathcal{L}^{P \leftrightarrow I}$ and $\mathcal{L}^{P \leftrightarrow T}$ (each has the similar formulation of Equation~\ref{eq_clip_loss}).
In this case, the optimization objective becomes:
\begin{equation}
\label{eq:naive_align}
\mathcal{L}^{P \leftrightarrow I}(\bm{x}_{i}^{P},\bm{x}_{i}^{I}) + 
\mathcal{L}^{P \leftrightarrow T}(\bm{x}_{i}^{P},\bm{x}_{i}^{T})
\end{equation} 

\paragraph{Revisiting Training Recipe.}
It is known to the 3D community that a well-tuned training recipe can lead to a dramatic performance boost~\cite{pointnext}. 
Yet, despite its impressively promising performance, the training recipe of the image-text-3D contrastive learning paradigm is underexplored.  
Thus, before diving deep into our method, we first revisit the training recipe of ULIP~\cite{ulip} and OpenShape~\cite{openshape}, identifying useful changes, as listed in Table \ref{tab:recipe}:

\begin{itemize}[leftmargin=*,itemsep=.5ex]
    \item \textbf{Batchsize.} Contrastive learning benefits significantly from a large batch size~\cite{openclip,openai_clip}. 
    Nevertheless, the state-of-the-art model~\cite{ulip} still adopts a small batchsize of 64.  
    We scale the batchsize and note a medium 2k strikes a good trade-off between different datasets. 
    \item \textbf{Learning rate schedule.} 
    Unlike ULIP, OpenShape adopts the step learning rate decay schedule without warmup. 
    We adopt the cosine learning rate schedule as CLIP and find it leads to clear improvement.
    \item \textbf{Exponential moving average.} During training, we observe the model performance steadily increase on the synthetic Objaverse-LVIS dataset, while fluctuating drastically on the real-scanned ScanObjectNN dataset, presumably due to the domain gap. 
    We employ Exponential Moving Average (EMA)~\cite{mean_teacher} to alleviate the fluctuation issue to stabilize training.
    \item \textbf{Separate temparature.} 
    Features from different modalities may have different distributions. Prior works~\cite{ulip,openshape} use a shared temperature parameter $\tau$~\cite{wu2018unsupervised} to control the concentration level of multi-modal features.
    We hereby use separate temperature parameters for each modality. 
\end{itemize}

\input{experiments/tab_bag_of_tricks_summary}

Together, as shown in Table \ref{tab:abl_bag_of_tricks}, our enhanced recipe substantially boosts the top-1 accuracy of OpenShape baseline by 3.3\%, 3.6\%, and 1.9\% on Objaverse-LVIS, ScanObjectNN, and ModelNet40, respectively. 
Next, we introduce our proposed \ourmethod~for 3D contrastive learning.

\begin{figure*}[!t]
  \centering
  \includegraphics[width=0.95\linewidth]{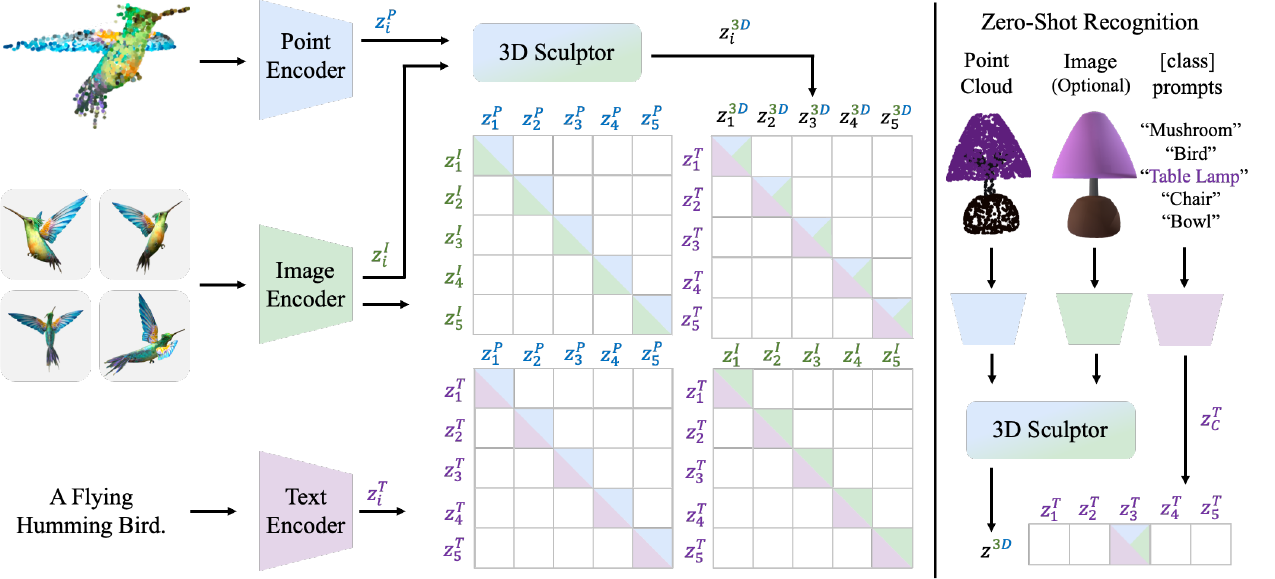}
   \caption{Summary of our \ourmethod~framework. \ourmethod~first extracts the representation of input triplets (images, text, point cloud) from a pre-trained vision-language model (\eg, CLIP) and a 3D encoder (\eg, Point-BERT) with corresponding projection heads. 
   Then, the image and point cloud features go through a 3D sculptor to obtain the 3D object-level features, serving as complementary representations. 
   The contrastive losses are applied to align features among three modalities (image-text-3D) and 3D representation to text.}
   \label{fig:method}
\end{figure*}
\subsection{3D-Text Alignment}
\label{subsec:joint_modal}
Point cloud and 2D images are known to encode different yet complementary cues: point cloud better captures depth and geometry information, while images focus on catching dense semantic information~\cite{bai2022transfusion,liu2023bevfusion,chen2023bevdistill,Wang2023distillbev}.
Meanwhile, in the 3D world, a single-view image only contains partial information captured from a specific camera pose with an angle.
Instead, multi-view, a prominent property in 3D representation, has demonstrated promising effectiveness in 3D understanding tasks~\cite{MVConv,jaritz2019multi,hamdi2021mvtn,hamdi2023voint}. 
Though previous works~\cite{ulip,ulip2,openshape} render images from multiple viewpoints of the same point cloud when creating the data triplets, they merely sample one image from the rendered multi-view images when extracting the image features, which inherently encode only partial facets of the 3D object.
\yp{Overall, both point cloud and multi-view images are two different perspectives from which a 3D object is treated.}
To this end, we introduce a simple yet effective \yp{3D-text alignment approach for contrastive language-image-3D pre-training}, which constructs a new 3D object-level representation by aggregating the respective features extracted from multi-view RGB images and point cloud modalities.
On top of the contrast between the conventional tri-modal features with each other, the joint representation will also be aligned with text features via a 3D-text contrastive loss.

\paragraph{Sculpting Holistic 3D Representation.}
Given batched data triplets $\{\bm{x}_i=(\bm{x}_{i}^{I}, \bm{x}_{i}^{T}, \bm{x}_{i}^{P})\}_{i=1}^{N}$ and image-text-3D models ($f^I$, $f^T$, $f^P$), the corresponding features are denoted as $\mathbb{R}^{D}$ vectors $(\bm{z}_{i}^{I}$, $\bm{z}_{i}^{T}$, $\bm{z}_{i}^{P}$), respectively.
With $M$ multi-view images $\bm{x}_i^{I}=\{\bm{x}_{(i,j)}^{I}\}_{j=1}^{M}$, which corresponds to the text description $\bm{x}_{i}^{T}$ and point cloud $\bm{x}_{i}^{P}$, we replace the single-view image feature with the fusion of individual image features $\bm{z}_{(i,j)}^{I}$ extracted from images $\bm{x}_{(i,j)}^{I}$.
We first construct the multi-view image feature $\bm{z}_{(i,j)}^{I}$ by fusing the features $\bm{z}_{(i,j)}^{I}$ of every single view $\bm{x}_{(i,j)}^{I}$:
\begin{equation}
    \bm{z}_{i}^{I} = g^{MV}(\{\bm{z}_{(i,j)}^{I}\}_{j=1}^{M})
\end{equation}
\noindent where $g^{MV}$ is the multi-view fusion function (\textit{e.g.}, view-pooling, maxpooling, or MLP) for comprehensive RGB representation modeling. 

To model a holistic 3D object-level representation, we concatenate the image features and point cloud features (\ie, $\mathrm{concat}(\bm{z}_{i}^{I}, \bm{z}_{i}^{P}) \in \mathbb{R}^{2 \times D}$), and use a 3D sculptor $g^{3D}$ (a fully connected layer) to project the joint representation as follows:
\begin{equation}
    \bm{z}_i^{3D} = g^{3D}(\mathrm{concat}(\bm{z}_{i}^{I}, \bm{z}_{i}^{P}))
\end{equation} 

\paragraph{Training Objectives.}
With a holistic 3D representation $\bm{z}_i^{3D}$ and text features $\bm{z}_{i}^{I}$, the 3D-text contrastive term is:

\begin{equation}
\label{eq:modal_joint_loss}
\mathcal{L}^{3D \leftrightarrow T}(\bm{x}_{i}^{I}, \bm{x}_{i}^{P},\bm{x}_{i}^{T}) = -\frac{1}{2} (l^{3D \rightarrow T} + l^{T \rightarrow 3D})
\end{equation}

where $l^{3D \rightarrow T}$ and $l^{T \rightarrow 3D}$ have similar formulation as shown in Equation~\ref{eq:logit1} and Equation~\ref{eq:logit2} with input $\bm{z}_i^{3D}$ and $\bm{z}_{i}^{I}$. 
In this case, the overall objective becomes:
\begin{equation}
\label{eq:total_loss}
\begin{aligned}
\mathcal{L}^{\mathrm{All}}(\bm{x}_i) = \mathcal{L}^{3D \leftrightarrow T} + \mathcal{L}^{P \leftrightarrow I} + \mathcal{L}^{P \leftrightarrow T} + \mathcal{L}^{I \leftrightarrow T}
\end{aligned}
\end{equation}

where $\{\bm{x}_i=(\bm{x}_{i}^{I}, \bm{x}_{i}^{T}, \bm{x}_{i}^{P})\}_{i=1}^{N}$ is the input image-text-point cloud data triplet.
Note that we keep the conventional point cloud to text loss $\mathcal{L}^{P \leftrightarrow T}$, enabling the model to make predictions solely based on 3D input even when corresponding images are unavailable \cite{scanobjectnn,modelnet40}.
Additionally, different from ULIP and OpenShape, we retain the CLIP loss $\mathcal{L}^{I \leftrightarrow T}$ with an additional learnable projection head upon the frozen CLIP encoder.

\input{experiments/tab_zero_shot_main_results}

\paragraph{Zero-Shot Inference.}
The texts of class labels in the downstream tasks are used to connect to the learned 3D representation from the point cloud encoder, enabling the ability of zero-shot recognition.
Specifically, the $C$ class text features $z^T_{C} \in \mathbb{R}^{C\times D} $ are obtained by inputting the class label to the text encoder with prompt engineering.
Then, for single and mixture modality inference, given the trained 3D sculptor $g^{3D}$ and extracted image $\bm{z}^I_i$, point cloud $\bm{z}^P_i$ features, the logits $y_i^{3D}$, $y_i^{P}$, $y_i^{I}$ between the 3D object and texts are calculated in different ways as follows:
\begin{equation}
\label{eq:2_to_1_inference}
\begin{aligned}
&y_i^{3D} = g^{3D}(\mathrm{concat}(\bm{z}_{i}^{I}, \bm{z}_{i}^{P})) \cdot z^T_{C}\ , \\
&y_i^{P} = \bm{z}_{i}^{P} \cdot z^T_{C}\ , y_i^{I} = \bm{z}_{i}^{I} \cdot z^T_{C}
\end{aligned}
\end{equation}

Note that our \ourmethod~also flexibly supports single-modality zero-shot inference, \textit{i.e.}, utilizing $y_i^{P}$ for point cloud-to-text~\cite{ulip, ulip2, openshape}) or $y_i^{I}$ for image-to-text~\cite{zhang2022pointclip, zhu2022pointclipv2}.

\section{Experiments}
\label{sec:experiments} 
We first introduce our experimental setup in Section~\ref{sec:exp_setup}.
Then, we compare previous state-of-the-art methods in Section~\ref{subsec:main_results}.
We also conduct a series of analyses on the key components (Section~\ref{subsec:abl_studies}), including the improved training strategies, contrastive loss, multi-view, and effect of inference ways.
Additionally, 
our \ourmethod~can benefit cross-modal applications such as text to 3D object retrieval and point cloud captioning (Section~\ref{subsec:visualization}).

\subsection{Experimental Setup}
\label{sec:exp_setup}
\paragraph{Pre-training datasets.}
Following OpenShape~\cite{openshape}, the full pre-training dataset (denoted as ``Ensemble") contains four pieces: ShapeNet~\cite{shapenet}, 3D-FUTURE~\cite{3d-future}, ABO~\cite{abo}) and Objaverse~\cite{objaverse}. 
The point cloud is obtained by sampling 10,000 points from the mesh surface and the color is interpolated based on the mesh textures. 
The images are rendered from 12 preset camera poses that cover the whole object uniformly.
Then, the paired texts are generated by BLIP~\cite{blip, blipv2} and Azure cognition services with GPT4~\cite{GPT4} to filter out noisy text. 
In addition, we verify the 3D open-world understanding ability of our method trained by the ShapeNet dataset only and the ensembled dataset except for the LVIS~\cite{lvis} categories (denoted as ``Ensemble (No LVIS)"), which have fewer categories in training data and constitute a more challenging scenario.

\input{experiments/tab_abl_bag_of_tricks}
\input{experiments/tab_abl_loss}

\paragraph{Down-stream datasets.}
Three datasets are used for evaluating zero-shot point cloud recognition:
\begin{itemize}[leftmargin=*,itemsep=.5ex]
    \item ModelNet40~\cite{modelnet40} is a synthetic dataset comprising 3D CAD models, including 9,843 training samples and 2,468 testing samples, distributed across 40 categories.

    \item ScanObjectNN~\cite{scanobjectnn} is a dataset composed of 3D objects acquired through real-world scanning techniques, encompassing a total of 2,902 objects that are systematically categorized into 15 distinct categories. 
We follow~\cite{ulip, ulip2, openshape} and use the variants provided by~\cite{point-bert} in our experiments. 

\item Objaverse-LVIS, an annotated subset of the Objaverse~\cite{objaverse}, incorporates a corpus of 46,832 shapes originating from 1,156 categories in LVIS dataset~\cite{lvis}.
\end{itemize}

\paragraph{Implementation details.}
We implement our approach in PyTorch~\cite{pytorch} and train the models on a server with 8 NVIDIA A5000 GPUs with a batch size of 2048. 
We train the model for 200 epochs with the AdamW~\cite{AdamW} optimizer, a warmup epoch of 10, and a cosine learning rate decay schedule~\cite{cosine_lr}. 
The base learning rate is set to 1e-3, based on the linear learning rate scaling rule~\cite{linear_lr_law}: $lr = base\_lr \times$ batchsize / 256. The EMA factor is set to 0.9995. 
Following~\citet{openshape}, OpenCLIP ViT-bigG-14~\cite{openclip} is adopted as the pretrained CLIP model.

\subsection{Main Results}
\label{subsec:main_results}
In Table~\ref{tab:zero_shot_main_results}, we compare the performance of our \ourmethod~with state-of-the-arts across two representative encoders, SparseConv~\cite{sparseconv} and PointBERT~\cite{point-bert}; three different training set, ``ShapeNet", ``Ensemble (No LVIS)", and ``Ensemble"; and three popular 3D recognition benchmarks, Objaverse-LVIS, ScanObjectNN, and ModelNet40. 

We observe that our \ourmethod~consistently exhibits superior performance on different scales of the dataset (From ``ShapeNet" to ``Ensemble" and types of 3D encoders (SparseConv and PointBERT)). 
Specifically, on the challenging long-tailed benchmark Objaverse-LVIS, \ourmethod~greatly improves the zero-shot Top1 accuracy from 46.8\% of OpenShape to \textbf{52.5\%} with PointBERT encoder and ``Ensemble'' training data.
In addition, when tested on the ScanObjectNN dataset that comprises scanned points of real objects and thus a bigger domain gap~\cite{scanobjectnn}, our \ourmethod~also achieves a significant performance boost of \textbf{6.4\%} (58.6\% \vs 52.2\%).
These results altogether validate the effectiveness of our proposed \ourmethod, demonstrating a more powerful open-world 3D understanding ability. 

\subsection{Ablation Studies}
\label{subsec:abl_studies}

\paragraph{Improved training recipe.}
We show the effect of improved training strategies in Table~\ref{tab:abl_bag_of_tricks}. 
The separate temperatures obtain a notable performance improvement (+ 0.8\% Top1 on ScanObjectNN), indicating the necessity of using separate dynamic scales of logits.
A larger batchsize benefits image-text-3D contrastive pre-training, significantly increasing 1.2\%/0.7\% Top1 accuracy on Objaverse-LVIS and ScanObjectNN.   
We observe a similar effect of the cosine learning rate schedule with warmup, achieving 48.5\% Top1 and 36.0\% Top1-C on Objaverse-LVIS without any additional training cost. 
Lastly, the exponential moving average update brings consistent improvement, especially on the ScanObjectNN (+ 1.5\%) with a larger domain gap.

\begin{figure*}[t!]
  \centering
  \includegraphics[width=1.0\linewidth]{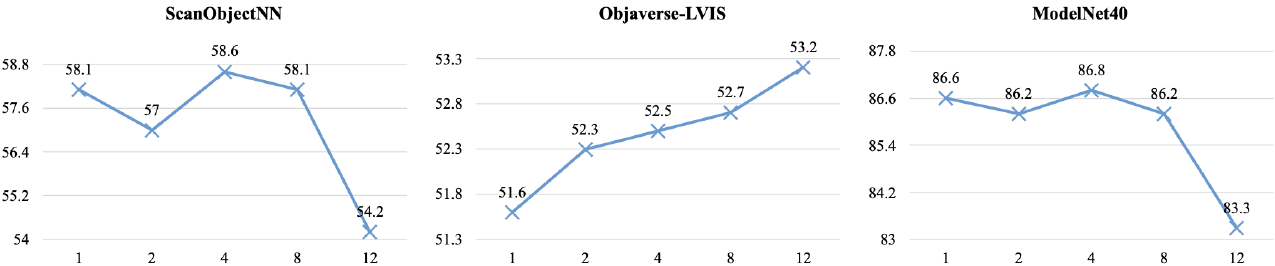}
   \caption{Analysis of the number in the multi-view mechanism.
  We report the Top 1 Accuracy results from \textbf{1} to \textbf{12} views.
   }
   \label{fig:abl_multi-view}
\end{figure*}
\input{experiments/tab_abl_g-mv}

\paragraph{\ourmethod~component.}
In Table~\ref{tab:abl_multi_modal_contrastive}, we analyze the effect of each critical component in \ourmethod. 
Interestingly, we find that the image-text alignment alone even leads to worse performance compared to the baseline (decreasing from 49.8\% to 48.7\% on Objaverse-LVIS and 86.1\% to 84.7\% on ModelNet40), which potentially hurts the alignment effectiveness on $\mathcal{L}^{P \leftrightarrow I}$ and $\mathcal{L}^{P \leftrightarrow T}$.  
By contrast, our proposed image-3D to text joint alignment loss $\mathcal{L}^{3D \leftrightarrow T}$ itself brings a considerable performance boost of at least 1.8\% on all three datasets, and combining $\mathcal{L}^{I \leftrightarrow T}$ leads to further improvement. 
This clearly shows the paramount importance of aggregating complimentary useful cues in contrastive learning with image, point cloud, and text. 
Moreover, we adopt multi-view images to construct a more comprehensive representation of the 3D object on image modality and result in a further improvement of 0.9\% and 0.5\% Top1 on Objaverse-LVIS and ScanObjectNN, suggesting the importance of considering the holism of 3D objects on cross-modal alignment.

\paragraph{Multi-view component.}
We next analyze the effect of fusion function $g^{MV}$ (Table~\ref{tab:abl_g-mv_right}) and the number of views (Figure~\ref{fig:abl_multi-view}) used during the pre-training.
For $g^{MV}$, compared with the max poling operation, we observe that simply adopting view-pooling achieves promising improvement (52.5\% v.s. 52.1\% Top1 on Objaverse-LVIS). 
Adding an additional fully connected layer (FC) after the pooling operation may boost the performance on in-distribution Objaverse-LVIS (+ 0.2\% Top1) while severely lowering the generalization ability on ScanObjectNN (- 6.2\% Top1). 
Since the image modality is only accessible when testing on Objaverse-LVIS, increasing the number of views during training obtains a consistent improvement (from 51.6\% to 53.2\% Top1 and 38.2\% to 39.5\% Top1-C) but may slightly hurt the performance on ScanObjectNN (decreasing 0.5\% Top1 when increasing the number of views from 4 to 8).
To keep a trade-off between datasets, we choose the view-pooling as $g^{MV}$ and view amount $M=4$ by default.
\input{experiments/tab_abl_mm_infer}

\begin{figure*}[t!]
  \centering
  \includegraphics[width=0.9\linewidth]{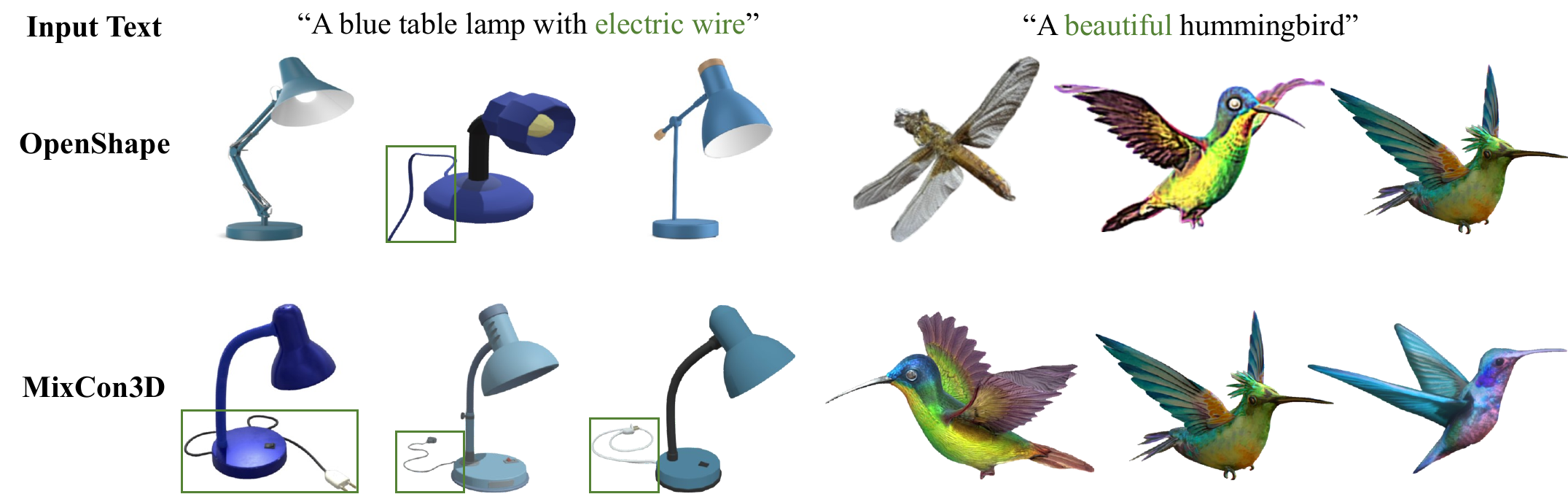}
  \vspace{-1em}
   \caption{\textbf{Text to 3D object retrieval comparisons.}
   The input text and the first three retrieved 3D objects are listed.
   }
   \label{fig:vis_text_3D_retrival}
\end{figure*}

\paragraph{Multi-modal Inference.}
The introduction of joint alignment and multi-view images leads to a lot of inference options. 
For instance, whether we should use point cloud input alone, or combine point cloud with image input. 
Also, it is necessary to decide whether to apply single-view or multi-view images for complete coverage.
We ablate a series of inference ways that aggregate different representations and show the results in Table~\ref{tab:abl_mm_infer}. 
Even with multi-view images, simply using the point cloud (50.4\% Top1) or image modality (51.6\% Top1) obtains sub-optimal solutions (compared to 52.5\% Top1 that uses modality fusion) since both only cover partial information of a 3D instance.
As can be seen, the way of point cloud and image representation fusion, plus multi-view image feature extraction (achieving 52.5\% Top1 and 38.8\% Top1-C), surpasses all other options by a clear margin, underpinning the significance of knowledge aggregation from different representations.

\begin{figure*}[t]
  \centering
  \includegraphics[width=0.9\linewidth]{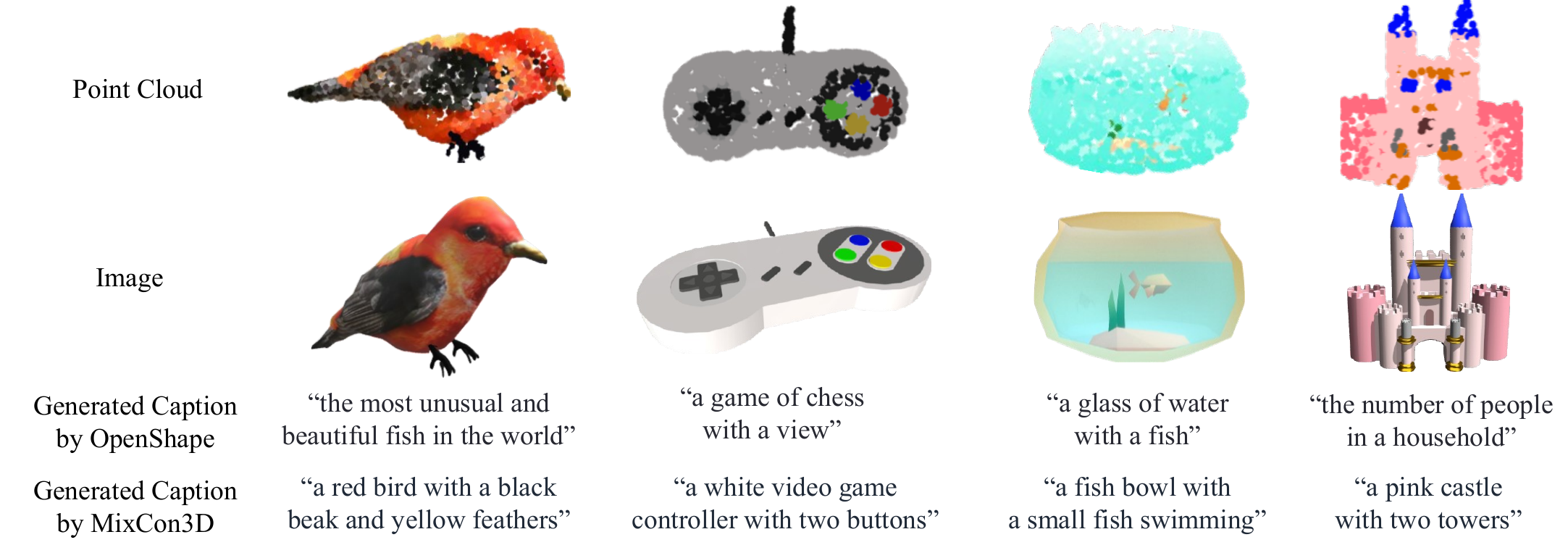}
    \vspace{-1em}
   \caption{\textbf{Point cloud captioning comparisons.}
   In each row, we list the input point cloud, corresponding images, and generated captions. 
   }
   \label{fig:vis_pc_cap}
\end{figure*}

\subsection{Cross-modal Applications}
\label{subsec:visualization}

To test how well the point cloud representation of our \ourmethod~is aligned with CLIP pre-trained representations, we conduct qualitative studies on the following cross-modal tasks, following the practice in \citet{openshape}. 

\paragraph{Text to 3D object retrieval.}
We use cosine similarity between text embeddings of a specific input and 3D shape embeddings from the ensembled dataset as the ranking metric. We compare the retrieval result of our \ourmethod~with that of OpenShape. As shown in Figure~\ref{fig:vis_text_3D_retrival}, our \ourmethod~can capture more comprehensive feature representation, \textit{e.g.}, allowing for more accurate indexing such as the hummingbird and fine-grained retrieval in situations where the ``lamp" is required to have an ``electric wire".

\paragraph{Point cloud captioning.}
We feed the 3D shape embeddings of our \ourmethod~into an off-the-shelf image captioning model ClipCap~\cite{mokady2021clipcap} and compare the results with that of OpenShape. 
As can be observed in Figure~\ref{fig:vis_pc_cap}, our \ourmethod~facilitates off-shelf models to generate more accurate and comprehensive captions, indicating that our method can better map the point cloud feature to the pre-aligned image-text feature space.

\section{Conclusion}
\label{sec:conclusion}
In this paper, we present \ourmethod, a simple yet effective image-text-3D contrastive learning approach, which synergizes multi-modal joint alignment and multi-view representations for better open-world 3D understanding capability.
Specifically, we propose constructing a simple yet effective 3D-text alignment training scheme and capitalizing on the features accumulated from multi-view images for a holistic 3D object-level representation.
In addition, we provide the first detailed training guideline in the field of contrastive language-image-3D pre-training.  
Together with the improved training pipeline, \ourmethod~achieves superior performance on a wide range of 3D recognition benchmarks and facilitates downstream cross-modal applications such as text-to-3D object retrieval and point cloud captioning. 
We hope our work could encourage more research endeavors to build the next-generation open-world 3D model. 

\section*{Acknowledgement}
This work is partially supported by TPU Research Cloud (TRC) program and Google Cloud Research Credits program.

{
    \small
    \bibliographystyle{ieeenat_fullname}
    \bibliography{main}
}


\clearpage

\input{supp_camera_ready}

\end{document}

%% file: preamble.tex
\usepackage[dvipsnames]{xcolor}
\newcommand{\red}[1]{{\color{red}#1}}


%% file: experiments/tab_bag_of_tricks_summary.tex
\begin{table}[t]
  \centering
  \setlength{\tabcolsep}{1.5pt}
\adjustbox{width=0.47\textwidth}{
  \begin{tabular}{c|cccccccc}
    \toprule
    Method           & \begin{tabular}[c]{@{}c}Temperature \\Parameter\end{tabular} & Batchsize & \begin{tabular}[c]{@{}c}Learning \\ Rate Schedule\end{tabular} & \begin{tabular}[c]{@{}c}Warm \\Up\end{tabular} & EMA\\
    \midrule
    ULIP       & Share & 64     & Cosine Decay & \Checkmark & \XSolidBrush \\
    OpenShape  & Share & 200    & Step Decay  & \XSolidBrush & \XSolidBrush \\
    Improved Recipe & Separate & $\sim$2k  & Cosine Decay & \Checkmark & \Checkmark\\
    \bottomrule
  \end{tabular}}
  \caption{The summary and comparisons between the baseline and our improved training recipe.}
\label{tab:recipe}
\end{table}

%% file: experiments/tab_zero_shot_main_results.tex
\begin{table*}[t]
\scriptsize
  \setlength{\tabcolsep}{1.8pt}
  \centering
  \adjustbox{width=\textwidth}{
    \begin{tabular}{l|c|c|cccc|cccc|cccc}
    \toprule
    \multirow{2}[1]{1.0cm}{Method} & \multirow{2}[1]{*}{Encoder} & \multirow{2}[1]{*}{\begin{tabular}[c]{@{}c}Training \\ data\end{tabular}}  & \multicolumn{4}{c|}{Objaverse-LVIS} & \multicolumn{4}{c|}{ScanObjectNN} & \multicolumn{4}{c}{ModelNet40} \\
\cmidrule{4-15}          
    & &  & Top1  & Top1-C & Top3  & Top5  & Top1 & Top1-C & Top3  & Top5  & Top1 & Top1-C & Top3  & Top5 \\
    \midrule
    PointCLIP~\cite{zhang2022pointclip} & - & \multirow{2}[1]{*}{\begin{tabular}[c]{@{}c}Depth \\ inference\end{tabular}} & 1.9   & - & 4.1   & 5.8   & 10.5 & - & 20.8      & 30.6 & 19.3 & - & 28.6      & 34.8 \\
    PointCLIP v2~\cite{zhu2022pointclipv2} & -&     & 4.7  & - & 9.5      & 12.9  & 42.2 & - & 63.3      & 74.5 & 63.6 & - & 77.9    & 85.0  \\
    \midrule
    ReCon~\cite{qi2023recon} & - & \multirow{8}[2]{*}{ShapeNet} & 1.1   & -  & 2.7      & 3.7      & 42.3   & - & 62.5      & 75.6 & 61.2 & - & 73.9      & 78.1      \\
    CG3D~\cite{clip_goes_3d}  & - &       & 5.0    & - & 9.5       & 11.6   & 42.5  &  -   & 57.3      & 60.8 & 48.7  & -  & 60.7      & 66.5    \\
    CLIP2Point~\cite{clip2point} & - &       & 2.7    &  - & 5.8     & 7.9      & 25.5  &  -  & 44.6      & 59.4   & 49.5 & - & 71.3 & 81.2     \\
     ULIP~\cite{ulip} & PointBERT &       & 6.2  & - & 13.6      & 17.9 & 51.5 & - & 71.1      & 80.2 & 60.4 & - & 79.0      & 84.4    \\
    OpenShape~\cite{openshape} & SparseConv & & 11.6 & - & 21.8 & 27.1 & 52.7 & - & 72.7 & \textbf{83.6} & 72.9 & - & 87.2 & 93.0\\
    \ourmethod~\textbf{(Ours)}& SparseConv & & \textbf{23.5} & \textbf{17.5} & \textbf{40.2} & \textbf{47.1} & \textbf{54.4} & \textbf{56.1} & \textbf{73.9} & 83.3 & \textbf{73.9} & \textbf{70.2} & \textbf{88.2} & \textbf{94.0} \\
    OpenShape~\cite{openshape} & PointBERT  & & 10.8 & - & 20.2 & 25.0 & 51.3 & - & 69.4 & 78.4 & 70.3 & - & 86.9 & 91.3 \\
    \ourmethod~\textbf{(Ours)}& PointBERT & & 22.3 & 16.2 & 37.5 & 44.3 & 52.6 & 52.1 & 69.9 & 78.7 & 72.6 & 68.2 & 87.1 & 91.3 \\
    \midrule
    ULIP~\cite{ulip} & PointBERT & \multirow{5}[1]{*}{\begin{tabular}[c]{@{}c}Ensemble \\ (No LVIS)\end{tabular}}  & 21.4 & - & 38.1      & 46.0     & 46.0   & - & 66.1      & 76.4 & 71.4 & - & 84.4      & 89.2      \\
    OpenShape~\cite{openshape} & SparseConv &  & 37.0 & - &   58.4    & 66.9  &    54.9  & - &  76.8     &  87.0 & 82.6  & - & 95.0    & 97.5  \\
    \ourmethod~\textbf{(Ours)} & SparseConv & & 45.7 & 33.5 & 67.0 & 73.2 & 56.5 & 60.5 & 77.8 & 87.5 & 83.3 & 82.4 & 95.6 & 97.6 \\
     OpenShape~\cite{openshape} & PointBERT & & 39.1 & - & 60.8 & 68.9 & 47.2 & - & 72.4 & 84.7 & 85.3 & - & 96.2 & 97.4 \\
    \ourmethod~\textbf{(Ours)} & PointBERT & & \textbf{47.5} & \textbf{34.6} & \textbf{69.0} & \textbf{76.2} & \textbf{57.7} & \textbf{61.5} & \textbf{80.7} & \textbf{89.8} & \textbf{87.3} & \textbf{86.7} & \textbf{96.8} & \textbf{98.1} \\
    \midrule
    ULIP~\cite{ulip} & PointBERT & \multirow{7}[2]{*}{Ensemble} & 26.8 & - & 44.8      & 52.6    & 51.6    & - & 72.5      & 82.3  & 75.1 & - & 88.1      & 93.2      \\
    OpenShape~\cite{openshape}  & SparseConv &       & 43.4 & - & 64.8 & 72.4 & 56.7 & - &   78.9    & 88.6 & 83.4 & - & 95.6 & 97.8 \\
    \ourmethod~\textbf{(Ours)} & SparseConv & & 47.3 & 35.0 & 68.7 & 76.1 & 57.1 & 61.2 & 79.2 & 88.9 & 83.9 & 83.2 & 95.9 & 98.0 \\
    OpenShape~\cite{openshape} & PointBERT & & 46.8 & 34.0 & 69.1 & 77.0 & 52.2  & 53.2 & 79.7 & 88.7 & 84.4 & 84.9 & 96.5 & 98.0 \\
    \ourmethod~\textbf{(Ours)} & PointBERT & & \textbf{52.5} & \textbf{38.8} & \textbf{74.5} & \textbf{81.2} & \textbf{58.6} & \textbf{62.3} & \textbf{80.3} & \textbf{89.2} & \textbf{86.8} & \textbf{86.8} & \textbf{96.9} & \textbf{98.3}\\
    \bottomrule
    \end{tabular}}%
    \caption{Comparison with state-of-the-art methods on three representative zero-shot 3D reognition benchmarks. ``Top1-C" means the top-1 class average accuracy. ``Encoder" denotes the point cloud encoder used in the framework. 
     "*" denotes the results we reproduce in public OpenShape datasets by corresponding methods.
  }
  \label{tab:zero_shot_main_results}%
\end{table*}%

%% file: experiments/tab_abl_bag_of_tricks.tex
\begin{table*}[t]
  \centering
  \vspace{-.5em}
  \renewcommand\tabcolsep{7pt}
  \resizebox{0.98\linewidth}{!}{
  \begin{tabular}{l|ccccccccccccccccccccccccccccc}
    \toprule
    \multirow{2}*{Improvements} & \multicolumn{4}{c}{\small Objaverse-LVIS}  & \multicolumn{4}{c}{\small ScanObjectNN} & \multicolumn{4}{c}{\small ModelNet40} \\
    \cmidrule(lr){2-5} \cmidrule(lr){6-9} \cmidrule(lr){10-13}
                 & Top1 & Top1-C & Top3 & Top5 & Top1 & Top1-C & Top3 & Top5 & Top1 & Top1-C & Top3 & Top5 \\
    \midrule
    Baseline    & 46.5   & 34.0   & 69.0 & 76.8 & 52.0   & 53.2 & 77.5 & 87.5 & 84.2 & 84.9 & 95.9 & 97.4 \\
    + Separate Temperature & 46.8 & 34.4 & 69.2 & 77.1 & 52.8 & 54.0 & 77.6 & 87.4 & 84.4 & 84.6 & 96.1 & 97.4\\
    + Large Batchsize & 48.0 & 35.3 & 70.1 & 77.4 & 53.5 & 55.5 & 78.0 & 87.7 & 84.8 & 85.3 & 96.4 & 97.7 \\
    + LR Schedule & 48.5 & 36.0 & 70.6 & 77.7 & 54.1 & 56.3 & 78.2 & 87.9 & 85.0 & 85.0 & 96.4 & 97.9\\
    + EMA & 49.8 & 36.9 & 71.7 & 78.7 & 55.6 & 58.9 & 79.3 & 88.6 & 86.1 & 86.2 & 96.8 & 98.3 \\
    \bottomrule
  \end{tabular}}
  \caption{Ablation studies for sequentially applying the improved training strategies for constructing a strong baseline on downstream zero-shot tasks. 
  The baseline denotes only using vanilla $\mathcal{L}^{P\leftrightarrow I}$ and $\mathcal{L}^{P\leftrightarrow T}$ with Point-BERT and OpenShape training recipe.}
  \label{tab:abl_bag_of_tricks}
\end{table*}

%% file: experiments/tab_abl_loss.tex
\begin{table*}[t]
  \centering
  \vspace{-.5em}
  \setlength{\tabcolsep}{7pt} 
  \adjustbox{width=\textwidth}{
  \begin{tabular}{ccc|cccccccccccccccccccccccccc}
    \toprule
    \multirow{2}*{$\mathcal{L}^{I \leftrightarrow T}$} & \multirow{2}*{$\mathcal{L}^{3D \leftrightarrow T}$} & \multirow{2}*{\makecell[c]{Multi- \\  View}} & \multicolumn{4}{c}{\small Objaverse-LVIS}  & \multicolumn{4}{c}{\small ScanObjectNN} & \multicolumn{4}{c}{\small ModelNet40}\\
    \cmidrule(lr){4-7} \cmidrule(lr){8-11} \cmidrule(lr){12-15}
    & &  & Top1 & Top1-C & Top3 & Top5 & Top1 & Top1-C & Top3 & Top5 & Top1 & Top1-C & Top3 & Top5\\
    \midrule
     \XSolidBrush & \XSolidBrush  & \XSolidBrush & \textbf{49.8} & \textbf{36.9} & \textbf{71.7} & \textbf{78.7} & \textbf{55.6} & 58.9 & \textbf{79.3} & \textbf{88.6} & \textbf{86.1} & \textbf{86.2} & \textbf{96.8} & \textbf{98.3} \\
    \checkmark & \XSolidBrush  &  \XSolidBrush & 48.7 & 36.2 & 70.4 & 77.7 & 55.4 & \textbf{59.7} & 75.8 & 85.6 & 84.7 & 84.8 & 96.6 & 97.9 \\
    \checkmark & \XSolidBrush  & \checkmark & 49.6 & 36.7 & 71.0 & 78.4 & 55.0 & 59.3 & 75.6 & 85.2 & 84.7 & 84.7 & 96.5 & 97.8\\
    \hline
    \specialrule{0em}{1pt}{1pt}
    \XSolidBrush  & \checkmark & \XSolidBrush & 51.0 & 37.8 & 73.2 & 79.5 & 57.9 & 61.4 & 79.8 & \textbf{89.3} & 86.5 & 86.4 & 96.6 & 98.0 \\
    \XSolidBrush  & \checkmark & \checkmark & 51.5 & 38.2 & 73.8 & 80.5 & 58.2 & 61.5 & 79.8 & 89.2 & 86.6 & 86.5 & 96.6 & 98.0 \\
    \checkmark  & \checkmark & \XSolidBrush & 51.6 & 38.2 & 73.7 & 80.6 & 58.1 & 61.9 & \textbf{80.3} & 89.2 & 86.6 & 86.6 & 96.4 & 98.1\\
    \checkmark  & \checkmark & \checkmark & \textbf{52.5} & \textbf{38.8} & \textbf{74.5} & \textbf{81.2} & \textbf{58.6} & \textbf{62.3} & \textbf{80.3} & 89.2 & \textbf{86.8} & \textbf{86.8} & \textbf{96.9} & \textbf{98.3}  \\
    \bottomrule
  \end{tabular}}
  \caption{The ablation studies of different optimization objectives in the proposed \ourmethod.}
  \label{tab:abl_multi_modal_contrastive}
\end{table*}

%% file: experiments/tab_abl_g-mv.tex
\begin{table}[t]
  \centering
  \setlength{\tabcolsep}{7pt}
  \resizebox{0.98\linewidth}{!}{
  \begin{tabular}{l|cccccccc}
    \toprule
    \multirow{2}*{\makecell[c]{Function $g^{MV}$}} & \multicolumn{2}{c}{\small O-LVIS} & \multicolumn{2}{c}{\small S-Object} \\
    \cmidrule(lr){2-3} \cmidrule(lr){4-5}
    & Top1 & Top1-C & Top1 & Top1-C\\
    \midrule
    - & 51.6 & 38.2 & 58.1 & 61.9\\
    \rowcolor{blue!8} View-pooling & 52.5 & 38.8 & \textbf{58.6} & \textbf{62.3} \\
    View-pooling + FC & \textbf{52.7} & \textbf{39.1} & 52.4 & 54.1\\
    Max pooling & 52.1 & 38.4 & 56.7 & 60.0\\
    Max pooling + FC & 51.6 & 38.0 & 55.8 & 58.7\\
    \bottomrule
  \end{tabular}}
  \caption{Ablation studies of the design of fusion function $g^{MV}$. 
  We report results on Objaverse-LVIS (O-LVIS) and ScanObjectNN (S-Object).}
  \label{tab:abl_g-mv_right}
\end{table}

%% file: experiments/tab_abl_mm_infer.tex
\begin{table}[t]
  \centering
  \setlength{\tabcolsep}{7pt}
  \resizebox{0.98\linewidth}{!}{
  \begin{tabular}{ccc|cccccc}
    \toprule
    \multirow{2}*{\makecell[c]{Point \\  Cloud}} & \multirow{2}*{Image} & \multirow{2}*{\makecell[c]{Multi- \\  View}} & \multicolumn{4}{c}{\small Objaverse-LVIS}\\
    \cmidrule(lr){4-7}
    &  &  & Top1 & Top1-C & Top3 & Top5\\
    \midrule
    \checkmark    & \XSolidBrush & - & 50.4 & 37.4 & 72.2 & 79.1\\
    \XSolidBrush  & \checkmark   & \XSolidBrush & 44.5 & 34.5 & 64.2 & 70.6\\
    \XSolidBrush  & \checkmark   & \checkmark & 51.9 & 38.5 & 73.1 & 79.4\\
    \checkmark    & \checkmark   & \XSolidBrush & 51.6 & 37.6 & 73.4 & 80.1 \\
    \rowcolor{blue!8} \checkmark    & \checkmark   & \checkmark & \textbf{52.5} & \textbf{38.8} & \textbf{74.5} & \textbf{81.2} \\
    \bottomrule
  \end{tabular}
  }
  \caption{Analysis of different types of 3D object-level representaton. 
  We report results on Objaverse-LVIS.}
  \label{tab:abl_mm_infer}
\end{table}

%% file: supp_camera_ready.tex


%
\definecolor{cvprblue}{rgb}{0.21,0.49,0.74}
\def\ie{\textit{i.e.}} 
\def\vs{\textit{vs.}} 
\def\eg{\textit{e.g.}}

\renewcommand\thetable{S\arabic{table}}  
\renewcommand\thefigure{S\arabic{figure}}
\renewcommand\thesection{S\arabic{section}}

\def\paperID{8626} 
\def\confName{CVPR}
\def\confYear{2024}




\section{Appendix}

\begin{table*}[ht]
\centering
\caption{Hyperparameters for scaling up PointBERT~\citep{point-bert}.}
\label{tab:pbertscl}
\begin{tabular}{ccccccc}
\toprule
\# Parameters & \# Layers & Width & \# Heads & MLP Dim & \# Patches & Patch Embed Dim  \\ 
\midrule
13.3M         & 6         & 512   & 8        & 1024    & 64         & 128                  \\
25.9M         & 12         & 512   & 8        & 1024    & 128         & 128                  \\
32.3M         & 12        & 512   & 8        & 1536    & 384        & 256                  \\
\bottomrule
\end{tabular}
\end{table*}

\subsection{Details of the 3D encoder}
\paragraph{The tokenization of point cloud.} 
We follow~\citet{point-bert} to partition the points into 512 point groups (sub-clouds), with a sub-cloud containing precisely 32 points. 
Then, a mini-PointNet~\citep{pointnet} is adopted to project those sub-clouds into point embeddings.

\paragraph{PointBERT backbone.}
Following OpenShape~\citep{openshape}, we scale up the Point-BERT~\citep{point-bert} model.
The hyperparameters for scaling up are shown in Table~\ref{tab:pbertscl}.

\input{experiments/tab_logit_appendix}
\input{experiments/tab_bs_appendix}

\subsection{More details and experimental results of the strong baseline}
We begin with the baseline developed by \citet{openshape} and use the ``Ensemble" dataset for training.
The temperature parameters for scaling the logit in two contrastive losses are unified, and the batchsize is 200.  
To simplify the analysis, we don't use the ``Hard Negative Mining" method utilized by OpenShape.

\paragraph{The setting of temperature for the contrastive loss.}
The temperature controls the range of logits in the softmax function used in the contrastive loss~\citep{openai_clip}.
We first follow the CLIP, which initializes the learnable temperature parameter to 14.28 and clamps the value if it exceeds 100.
In the image-text-3D alignment paradigm, the point cloud encoder is trained to align image and text modalities simultaneously. 
Intuitively, different modalities may have separately appropriate logit ranges.
To this end, we verify the effect of temperature settings (a \textit{unified} one used by two losses or two \textit{separate} ones, each used by a loss) in Table~\ref{tab:logit_abl} and choose ``Clamp+Separate" by default.

\paragraph{The effect of batchsize for different model sizes}.
We systematically investigate the effect of batchsize across model sizes in Table~\ref{tab:sc_bs_abl}.
From the results, increasing the model size or batchsize can obtain a better performance on Objaverse-LVIS whose distribution matches the training set very well.
However, the results of the other two datasets are barely satisfactory, indicating the model's generalization ability trained by ``Ensemble" dataset still has much room to improve.
Considering a good trade-off between datasets and training efficiency, we use a medium model size of ``25.9M" and batchsize of 2k for all the following ablation studies by default.

\paragraph{Hyperparameter analysis of EMA decay rate}.
We analyze the effect of the decay rate used in the Exponential-Moving-Average (EMA) update. 
From the results shown in Table~\ref{tab:ema_decay_abl}, choosing the decay rate from the range ``0.999" to ``0.9999" all yield promising results. 
Based on the results from three datasets, we choose ``0.9995" by default.

\input{experiments/tab_ema_decay_appendix}
\begin{figure*}[t]
  \centering
  \includegraphics[width=0.95\linewidth]{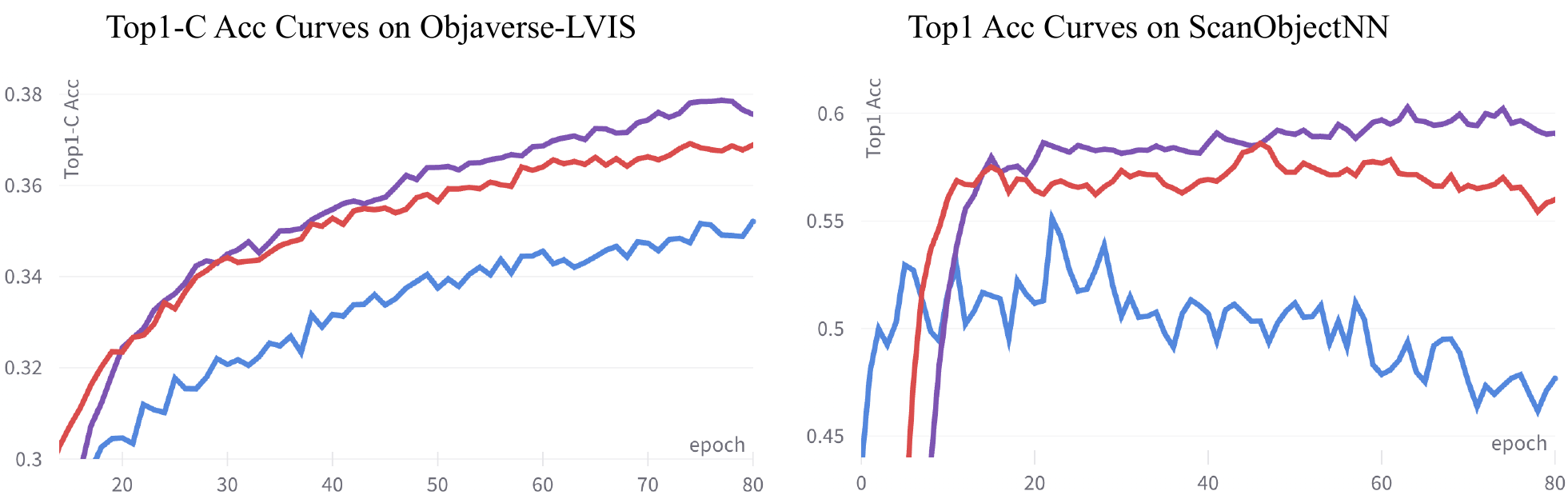}
   \caption{The zero-shot Top1 accuracy curve comparisons between the \blue{baseline},
   the improved \textcolor{red}{strong baseline} and our \textcolor{violet}{\ourmethod}.
   Our improved baseline can not only perform better on the Objaverse-LVIS benchmark 
   (the left sub-figure) but also stabilize the 
   generalization performance 
   (the right sub-figure).
   }
   \label{fig:training_recipe}
\end{figure*}

\paragraph{Training stability.}
Empirically, we observe that the model's test performance on the ScanObjectNN benchmark is unstable during training on the Objaverse dataset (the \blue{blue} curve in Figure~\ref{fig:training_recipe}).
Our improved baseline (the \red{red} curve) can significantly alleviate the training instability.
Meanwhile, our proposed \ourmethod~further boosts the performance for both the Objaverse-LVIS and ScanObjectNN.

\input{experiments/tab_mv_mm_joint_appendix}

\subsection{Additional Experimental Results}
\paragraph{Full results of $g^{MV}$ and view amounts.}
We list the full results of using various types of $g^{MV}$, and view amounts in Table~\ref{tab:g-mv_full_abl} and Table~\ref{tab:mv_full_abl}.
Using simple view-pooling as $g^{MV}$ obtains consistent improvement across three datasets. 
Adding an additional FC layer after the view-pooling or max pooling can enhance the Objaverse-LVIS performance while degrading the generalization ability on ScanObjectNN and ModelNet40.
Given the availability of the image modality in the Objaverse-LVIS testing scenario, an increase in the number of views during the training phase yields a consistent enhancement in performance. 
However, this increment marginally impairs the efficacy of the ScanObjectNN and ModelNet40.

\paragraph{A unified view of multi-modal inference}
We analyze various multi-modal feature ensemble methods under four views ($M=4$) in the main text.
To further analyze the combined impact of view amount and inference schemes, we perform in-depth analysis in Table~\ref{tab:mv-mm_joint_abl}, including individual modality inference (point cloud $y_{i}^P$ and image $y_{i}^P$) and modality ensemble inference (using $g^{MV}$ to obtain $y_{i}^{3D}$ and $y_{i}^P + y_{i}^I$).
From the results, the ensemble scheme of the point cloud and the image modalities significantly improves performance.
Moreover, benefitting from the large-scale pretrained CLIP model, the $y_{i}^P + y_{i}^I$ scheme further boosts the performance on Objaverse-LVIS when using multi-view images for inference.

\input{experiments/tab_g-mv_full_appendix}
\input{experiments/tab_multiiew_full_appendix}




%% file: experiments/tab_logit_appendix.tex
\begin{table*}[h]
  \centering
  \renewcommand\tabcolsep{3.2pt}
  \caption{The analysis of settings of temperature for the constrastive losses.}
  \resizebox{0.98\linewidth}{!}{
  \begin{tabular}{cc|cccccccccccccccccccccccccccc}
    \toprule
    \multirow{2}*{Clamp} & \multirow{2}*{{\makecell[c]{Temperature \\  Setting}}} & \multicolumn{4}{c}{\small Objaverse-LVIS}  & \multicolumn{4}{c}{\small ScanObjectNN} & \multicolumn{4}{c}{\small ModelNet40}\\
    \cmidrule(lr){3-6} \cmidrule(lr){7-10} \cmidrule(lr){11-14}
    &  & Top1 & Top1-C & Top3 & Top5 & Top1 & Top1-C & Top3 & Top5 & Top1 & Top1-C & Top3 & Top5\\
    \midrule
    \XSolidBrush &  Unified  & 46.5   & 34.0   & 69.0 & 76.8 & 52.0   & 53.2 & 77.5 & 87.5 & 84.2 & \textbf{84.9} & 95.9 & 97.4 \\
    \checkmark & Unified  & 46.5 & 34.1 & 69.0 & 76.8 & 52.2 & 53.3 & 77.5 & \textbf{87.7} & \textbf{84.4} & \textbf{84.9} & 96.0 & \textbf{97.6}\\
    \XSolidBrush & Separate  & 46.4 & 34.0 & 69.0 & 76.8 & 52.5 & 53.7 & 77.2 & 87.2 & 84.3 & 84.6 & 96.0 & 97.5\\
    \checkmark & Separate  & \textbf{46.8} & \textbf{34.4} & \textbf{69.2} & \textbf{77.1} & \textbf{52.8} & \textbf{54.0} & \textbf{77.6} & 87.4 & \textbf{84.4} & 84.6 & \textbf{96.1} & 97.4\\
    \bottomrule
  \end{tabular}}
  \label{tab:logit_abl}
\end{table*}

%% file: experiments/tab_bs_appendix.tex
\begin{table*}[b]
  \centering
  \renewcommand\tabcolsep{4pt}
  \caption{The analysis of batchsize across different model sizes.}
  \resizebox{0.98\linewidth}{!}{
  \begin{tabular}{cc|cccccccccccccccccccccccccccc}
    \toprule
    \multirow{2}*{Para.} & \multirow{2}*{Batchsize} & \multicolumn{4}{c}{\small Objaverse-LVIS}  & \multicolumn{4}{c}{\small ScanObjectNN} & \multicolumn{4}{c}{\small ModelNet40}\\
    \cmidrule(lr){3-6} \cmidrule(lr){7-10} \cmidrule(lr){11-14}
    &  & Top1 & Top1-C & Top3 & Top5 & Top1 & Top1-C & Top3 & Top5 & Top1 & Top1-C & Top3 & Top5\\
    \midrule
    \multirow{5}[1]{*}{32.3M}  & 256  & 47.0 & 34.9 & 69.2 & 76.9 & 50.2 & 52.6 & 77.9 & 87.5 & 85.0 & 85.2 & 96.4 & 97.4\\
     & 512  & 47.7 & 36.1 & 69.8 & 77.2 & 49.9 & 52.1 & 75.7 & 85.5 & \textbf{84.6} & \textbf{84.6} & 95.8 & 97.1\\
     & 1024 & 49.1 & 36.9 & 70.9 & 78.1 & 52.1 & \textbf{55.0} & 76.0 & \textbf{85.9} & 83.3 & 83.7 & \textbf{96.8}  & \textbf{98.3}\\
    & \textbf{2048} & \textbf{49.6} & 37.4 & \textbf{71.1} & \textbf{78.3}  & \textbf{53.2} & \textbf{55.0} & 75.3 & 85.7 & \textbf{84.6} & 83.7 & 95.2 & 96.9 \\
    & 4096 & \textbf{49.6} & \textbf{37.9} & 70.9 & 78.1  & 52.7 & 54.1 & \textbf{76.1} & 85.3 & 83.7 & 82.3 & 96.1 & 97.7\\
    \hline
    \multirow{5}[1]{*}{25.9M} & 256  & 46.8 & 34.4 & 69.2 & 77.1 & 52.8 & 54.0 & 77.6 & 87.4 & 84.4 & 84.6 & 96.1 & 97.4\\
     & 512  & 47.3 & 34.7 & 69.6 & 77.1 & 52.5 & 55.6 & 77.2 & 87.4 & 84.3 & 84.6 & 96.3 & \textbf{98.1}\\
     & 1024 & 47.8 & 35.0 & 69.9 & 77.2 & 52.9 & \textbf{56.2} & 77.7 & 87.5 & 84.4 & 85.3 & 96.3 & 98.0\\
     & \textbf{2048} & 48.0 & 35.3 & 70.1 & 77.4 & \textbf{53.5} & 55.5 & \textbf{78.0} & 87.7 & \textbf{84.8} & 85.3 & \textbf{96.4} & 97.7 \\
      & 4096 & \textbf{48.5} & \textbf{35.6} & \textbf{70.4} & \textbf{77.6} & 52.9 & 55.1 & 77.9 & \textbf{87.8} & 84.3 & \textbf{85.4} & 95.9 & 97.5\\
    \hline
    \multirow{3}[1]{*}{13.3M} & 512 & 45.2 & 33.7 & 66.7 & 74.5 & \textbf{54.7} & \textbf{56.6} & 77.3 & 87.0 & 83.7 & 83.7 & 94.7 & 96.8\\
    & \textbf{1024} & 45.8 & 34.3 & 67.1 & \textbf{74.8} & 54.2 & 56.4 & 76.3 & 86.4 & \textbf{85.2} & \textbf{84.0} & \textbf{95.6} & \textbf{97.7} \\
    & 2048 & \textbf{46.3} & \textbf{35.1} & \textbf{67.3} & \textbf{74.8} & 53.1 & 54.4 & \textbf{78.5} & \textbf{87.4} & 83.5 & 83.4 & 95.4 & 97.5 \\
    \bottomrule
  \end{tabular}}
  \label{tab:sc_bs_abl}
\end{table*}

%% file: experiments/tab_ema_decay_appendix.tex
\begin{table*}[t]
  \centering
  \renewcommand\tabcolsep{4pt}
  \caption{The analysis of settings of temperature for the constrastive losses.}
  \resizebox{0.98\linewidth}{!}{
  \begin{tabular}{l|cccccccccccccccccccccccccccc}
    \toprule
    \multirow{2}*{Decay Rate} & \multicolumn{4}{c}{\small Objaverse-LVIS}  & \multicolumn{4}{c}{\small ScanObjectNN} & \multicolumn{4}{c}{\small ModelNet40}\\
    \cmidrule(lr){2-5} \cmidrule(lr){6-9} \cmidrule(lr){10-13}
    & Top1 & Top1-C & Top3 & Top5 & Top1 & Top1-C & Top3 & Top5 & Top1 & Top1-C & Top3 & Top5\\
    \midrule
    \textit{w/o EMA} & 48.5 & 36.0 & 70.6 & 77.7 & 54.1 & 56.3 & 78.2 & 87.9 & 85.0 & 85.0 & 96.4 & 97.9  \\
    0.99     & 49.2 & 36.5 & 71.1 & 78.3 & 54.8 & 57.1 & 78.9 & 88.2 & 85.7 & 85.9 & 96.8 & \textbf{98.4}\\
    0.999    & 49.3 & 36.5 & 71.2 & 78.3 & 55.3 & 58.3 & 79.4 & 88.8 & \textbf{86.4} & \textbf{86.3} & \textbf{96.9} & \textbf{98.4} \\
    0.9995   & 49.8 & 36.9 & \textbf{71.7} & \textbf{78.7} & \textbf{55.6} & \textbf{58.9} & \textbf{79.3} & \textbf{88.6} & 86.1 & 86.2 & 96.8 & 98.3  \\
    0.9999   & \textbf{50.1} & \textbf{37.0} & 71.3 & 78.6 & 55.4 & 58.5 & 78.9 & 88.4 & 85.7 & 85.3 & 96.9 & 98.2 \\
    0.99999  & 0.3 & 0.1 & 0.5 & 1.1 & 17.1 & 11.2 & 25.2 & 42.6 & 5.3 & 5.4 & 13.5 & 21.6 \\
    \bottomrule
  \end{tabular}}
  \label{tab:ema_decay_abl}
\end{table*}

%% file: experiments/tab_mv_mm_joint_appendix.tex
\begin{table*}[thpb]
  \centering
  \renewcommand\tabcolsep{2pt}
  \caption{The ablations of inference schemes under different settings of views ($M$).}
  \resizebox{0.98\linewidth}{!}{
  \begin{tabular}{c|cccccccccccccccc}
    \toprule
    \multirow{2}*{{\makecell[c]{Inference \\Scheme}}} & \multicolumn{4}{c}{$M=1$}  & \multicolumn{4}{c}{$M=4$} & \multicolumn{4}{c}{$M=8$} & \multicolumn{4}{c}{$M=12$} \\
    \cmidrule(lr){2-5} \cmidrule(lr){6-9} \cmidrule(lr){10-13} \cmidrule(lr){14-17}
    & \small Top1 & \small Top1-C & \small Top3 & \small Top5 & \small Top1 & \small Top1-C & \small Top3 & \small Top5 & \small Top1 & \small Top1-C & \small Top3 & \small Top5 & \small Top1 & \small Top1-C & \small Top3 & \small Top5 \\
    \midrule
    $y_i^P$           & 51.1 & 37.9 & 73.2 & 80.0 & 50.4 & 37.4 & 72.2 & 79.1 & 51.1 & 38.4 & 73.1 & 79.8 & 51.5 & 39.4 & 73.7 & 80.5 \\
    $y_i^I$           & 45.1 & 34.6 & 64.3 & 70.8 & 51.9 & 38.5 & 73.1 & 79.4 & 52.0 & 41.1 & 73.1 & 79.5 & 52.5 & 41.5 & 73.8 & 80.1 \\
    $y_i^{3D}$     & \textbf{51.6} & \textbf{38.2} & \textbf{73.7} & \textbf{80.6} & 52.5 & 38.8 & 74.5 & 81.2 & 52.8 & 39.1 & 74.7 & 81.5 & 53.2 & 39.5 & 75.4 & 82.1\\
    $y_i^P + y_i^I$   & 51.2 & 37.8 & 73.1 & 79.6 & \textbf{53.8} & \textbf{40.9} & \textbf{75.5} & \textbf{81.9} & \textbf{54.8} & \textbf{43.1} & \textbf{76.3} & \textbf{82.7} & \textbf{55.3} & \textbf{43.8} & \textbf{77.1} & \textbf{83.4} \\
    \bottomrule
  \end{tabular}}
  \label{tab:mv-mm_joint_abl}
\end{table*}

%% file: experiments/tab_g-mv_full_appendix.tex
\begin{table*}[t]
  \centering
  \renewcommand\tabcolsep{4pt}
  \caption{The analysis of variants of $g^{MV}$.}
  \resizebox{0.98\linewidth}{!}{
  \begin{tabular}{l|cccccccccccccccccccccccccccc}
    \toprule
    \multirow{2}*{Function $g^{MV}$} & \multicolumn{4}{c}{\small Objaverse-LVIS}  & \multicolumn{4}{c}{\small ScanObjectNN} & \multicolumn{4}{c}{\small ModelNet40}\\
    \cmidrule(lr){2-5} \cmidrule(lr){6-9} \cmidrule(lr){10-13}
    & Top1 & Top1-C & Top3 & Top5 & Top1 & Top1-C & Top3 & Top5 & Top1 & Top1-C & Top3 & Top5\\
    \midrule
    - & 51.6 & 38.2 & 73.7 & 80.6 & 58.1 & 61.9 & \textbf{80.3} & \textbf{89.2} & 86.6 & 86.6 & 96.4 & 98.1  \\
    View-pooling      & 52.5 & 38.8 & 74.5 & 81.2 & \textbf{58.6} & \textbf{62.3} & \textbf{80.3} & \textbf{89.2} & \textbf{86.8} & \textbf{86.8} & \textbf{96.9} & \textbf{98.3} \\
    View-pooling + FC & \textbf{52.7} & \textbf{39.1} & \textbf{74.8} & \textbf{81.4} & 52.4 & 54.1 & 75.2 & 86.5 & 84.5 & 84.0 & 95.1 & 96.5\\
    Max pooling       & 52.1 & 38.4 & 74.1 & 80.4 & 56.7 & 60.0 & 79.3 & 89.1 & 85.9 & 85.6 & 96.9 & 98.1\\
    Max pooling + FC  & 51.6 & 38.0 & 73.2 & 80.3 & 55.8 & 58.7 & 77.1 & 87.6 & 85.2 & 85.6 & 96.0 & 97.6 \\
    \bottomrule
  \end{tabular}}
  \label{tab:g-mv_full_abl}
\end{table*}

%% file: experiments/tab_multiiew_full_appendix.tex
\begin{table*}[t]
  \centering
  \renewcommand\tabcolsep{4pt}
  \caption{Ablation studies for the amount ($M$) of the view.}
  \resizebox{0.98\linewidth}{!}{
  \begin{tabular}{c|cccccccccccccccccccccccccccc}
    \toprule
    \multirow{2}*{Multi-View} & \multicolumn{4}{c}{\small Objaverse-LVIS}  & \multicolumn{4}{c}{\small ScanObjectNN} & \multicolumn{4}{c}{\small ModelNet40}\\
    \cmidrule(lr){2-5} \cmidrule(lr){6-9} \cmidrule(lr){10-13}
    & Top1 & Top1-C & Top3 & Top5 & Top1 & Top1-C & Top3 & Top5 & Top1 & Top1-C & Top3 & Top5\\
    \midrule
    1 & 51.6 & 38.2 & 73.7 & 80.6 & 58.1 & 61.9 & \textbf{80.3} & \textbf{89.2} & 86.6 & 86.6 & 96.4 & 98.1  \\
    2   & 52.3 & 38.9 & 74.1 & 80.0 & 57.0 & 60.5 & 77.8 & 88.0 & 86.2 & 86.7 & 96.2 & 97.8\\
    4   & 52.5 & 38.8 & 74.5 & 81.2 & \textbf{58.6} & \textbf{62.3} & \textbf{80.3} & \textbf{89.2} & \textbf{86.8} & \textbf{86.8} & \textbf{96.9} & \textbf{98.3}\\
    8   & 52.7 & 39.3 & 74.7 & 81.7 & 58.1 & 61.7 & 78.9 & 88.5 & 86.2 & 85.5 & 96.8 & 98.1\\
    12  & \textbf{53.2} & \textbf{39.5} & \textbf{75.4} & \textbf{82.1} & 54.2 & 56.1 & 77.8 & 86.7 & 83.3 & 83.6 & 95.1 & 96.8 \\
    \bottomrule
  \end{tabular}}
  \label{tab:mv_full_abl}
\end{table*}